\documentclass[10pt,twocolumn,letterpaper]{article}

 \usepackage[pagenumbers]{cvpr} 
 \usepackage{multirow}
\definecolor{cvprblue}{rgb}{0.21,0.49,0.74}
\usepackage[pagebackref,breaklinks,colorlinks,allcolors=cvprblue]{hyperref}

\usepackage{float}
\usepackage[table]{xcolor}
\def\paperID{*****} 
\def\confName{CVPR}
\def\confYear{2026}

\title{VAG: Dual-Stream Video-Action Generation for Embodied Data Synthesis }

\author{Xiaolei Lang$^{1,2*}$ \, \, Yang Wang$^{1*}$ \, \, Yukun Zhou$^{1}$ \, \, Chaojun Ni$^{1,3}$ \\  Kerui Li$^{1,4}$ \, \, Jiagang Zhu$^{1}$ \, \, Tianze Liu$^{1}$ \, \, Jiajun Lv$^{2}$ \, \, Xingxing Zuo$^{5}$ \\ Yun Ye$^{1}$ \, \, Guan Huang$^{1}$ \, \, Xiaofeng Wang$^{1}$ \, \, Zheng Zhu$^{1\dagger}$ \\
$^{1}$GigaAI \, \, $^{2}$Zhejiang University \, \, $^{3}$Peking University \, \, \\$^{4}$ Institute of Automation, Chinese Academy of Sciences \\	$^{5}$Robotics Department, Mohamed bin Zayed University of Artificial Intelligence
}

\begin{document}
\twocolumn[{
\maketitle
\begin{figure}[H]
\hsize=\textwidth
\centering
\includegraphics[width=\textwidth]{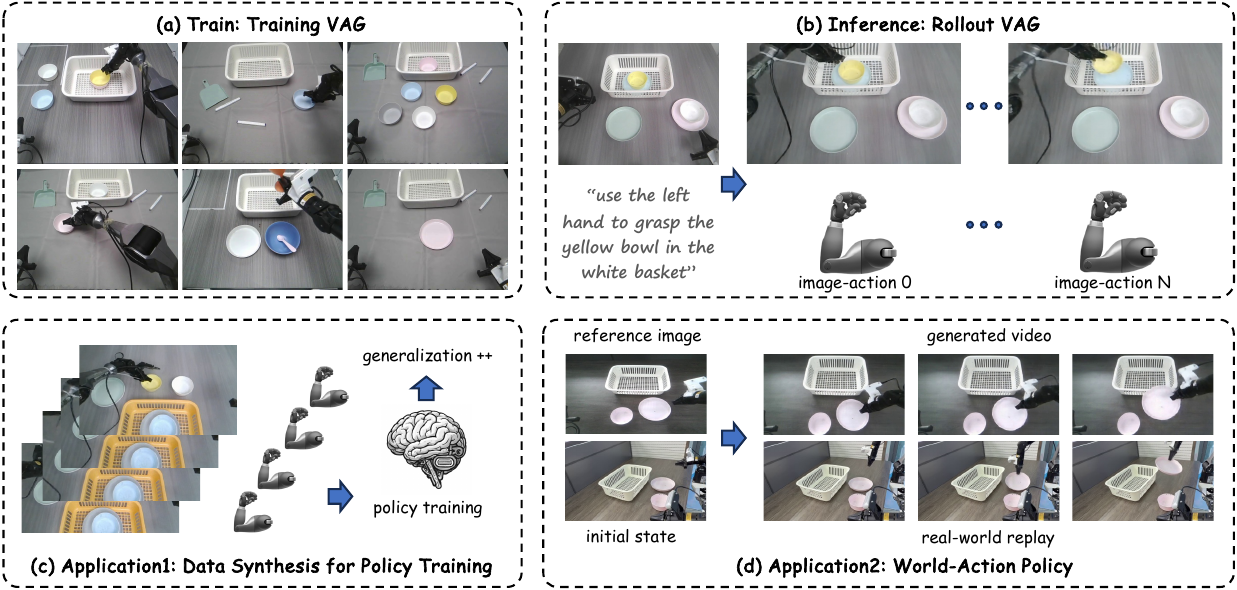}
\caption{\textbf{Illustration and capabilities of VAG.} (a) We train our dual-stream video-action generation model using teleoperated robot trajectories. (b) Given an initial frame and a language instruction, the model can synchronously generate aligned video–action data pairs. (c) The generated data can be used to train robot policies for improved generalization. (d) The actions generated by the model can also be applied for real-world robot replay. The successful task execution demonstrates that the model holds the potential to function as a policy.}
\label{fig:teaser}
\end{figure}
}]

\renewcommand{\thefootnote}{\fnsymbol{footnote}}
\footnotetext[1]{
These authors contributed equally to this work. 
}
\footnotetext[2]{Corresponding author: Zheng Zhu, zhengzhu@ieee.org.}

\begin{abstract}
Recent advances in robot foundation models trained on large-scale human teleoperation data have enabled robots to perform increasingly complex real-world tasks. However, scaling these systems remains difficult because collecting task-specific demonstrations is expensive and labor-intensive. Synthetic data, especially generated videos, offer a promising direction, but existing World Models (WMs) are not directly suitable for policy learning since they do not provide paired action trajectories. World-Action (WA) models partially address this by predicting actions with visual outputs, yet often lack strong video-action alignment, while two-stage pipelines that generate video first and then infer actions introduce inefficiency and error accumulation. To address these limitations, we propose VAG, a unified flow-matching-based dual-stream framework that jointly generates video and action under visual and language conditioning. By synchronizing denoising in both branches and using an adaptive 3D pooling mechanism to transfer compact global video context to the action branch, VAG improves cross-modal consistency during generation. Across both simulated and real-world settings, VAG produces aligned video-action pairs with competitive prediction quality, supports executable trajectory replay, and provides useful synthetic pretraining data that improves downstream policy generalization, indicating its potential as a practical world-action model for embodied data synthesis.
\end{abstract}    
\section{Introduction}
\label{sec:intro}

\begin{figure*}[t!]
    	\centering
    	\includegraphics[width=\textwidth]{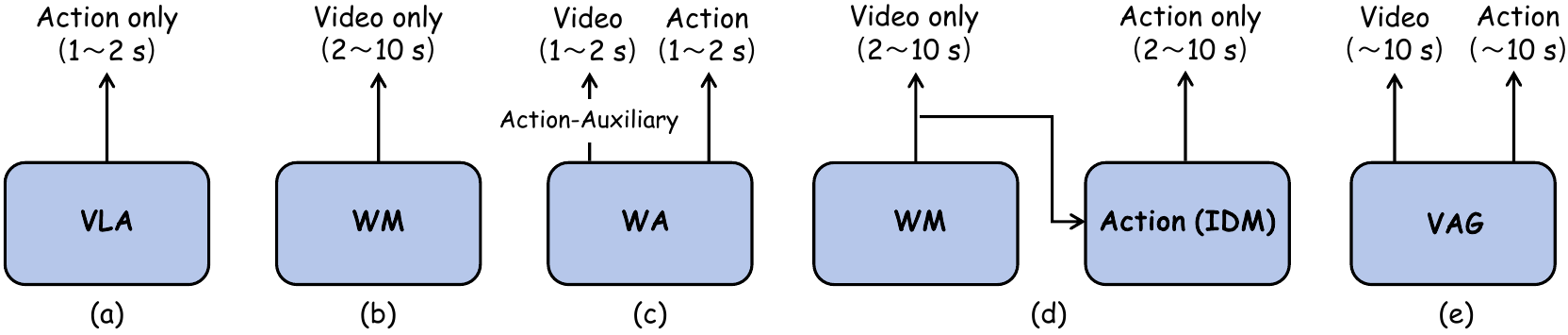}

    	\caption{\textbf{Architecture comparison of embodied models.} (a) Vision-Language-Action (VLA) models iteratively predict and execute actions, serving as a policy~\cite{pi_0}. (b) World models (WMs) generate rich visual rollouts spanning diverse scenarios but lack aligned action trajectories for direct policy learning~\cite{agarwal2025cosmos}. (c) World-Action (WA) models serve as a policy and enhance action prediction by incorporating video generation as an Action-Auxiliary signal~\cite{cen2025worldvla}. (d) World-Action (WA) models synthesize robot training data by combining world
        models and IDM~\cite{jang2025dreamgen}. (e) Our framework VAG specializes in joint and aligned video-action generation for embodied data synthesis.}
    \label{fig:intro}
\end{figure*}

Robot foundation models trained on large-scale human teleoperation data have demonstrated remarkable potential in enabling robotic systems to perform complex, dexterous tasks in the real world~\cite{pi_0, team2025gigabrain, kim24openvla, cheang2025gr3technicalreport, zhai2025igniting, intelligence2025pi05visionlanguageactionmodelopenworld, li2025cogvla, wang2025flowram, shukor2025smolvla, ni2026swiftvla, lin2025onetwovla}  Fig.~\ref{fig:intro}~(a). However, the need to manually collect teleoperation data for each new task or environment introduces significant cost and labor, becoming a major bottleneck to scalable robot learning. 

Fortunately, advancements in fields such as video and image synthesis~\cite{rombach2022high,humandreamer,OFFSET,HUD,PAIR,MEDIAN} offer promising alternatives. For instance, video generation can endlessly synthesize diverse embodied scenario videos, and recent progress in World Models (WMs) has significantly enhanced the realism and controllability of the generated videos~\cite{blattmann2023stable, wan2025, agarwal2025cosmos, chi2025wowworldomniscientworld, liu2025robotransfergeometryconsistentvideodiffusion, realmdreamer, Recondreamer-rl, humandreamerx, wang2025drivegen3d, zhao2025recondreamer, ni2025wonderfree, ni2025wonderturbo, zhao2025drivedreamer4d, team2025gigaworld}. Although producing rich visual information, they fail to directly facilitate policy learning, since the generated clips lack the paired action trajectories (Fig.~\ref{fig:intro}~(b)). 

To bridge this gap, World-Action (WA) models also predict actions for the following frames, instead of solely forecasting future observations~\cite{wu2023unleashing, cheang2024gr2generativevideolanguageactionmodel, cen2025worldvla, won2025dualstreamdiffusionworldmodelaugmented, li2025unified, bi2025motus, ye2026world, ye2026gigaworld, yuan2026fast}, as depicted in Fig.~\ref{fig:intro}~(c). Yet, they mainly focus on enhancing action prediction through predicted visual information, overlooking the alignment between the video and the action for policy-training-oriented data synthesis. In contrast, some approaches adopt a two-stage paradigm~\cite{jang2025dreamgen, tan2025anypos, li2026causal, rhoda2026dva}, where the video is generated first and then the actions are extracted from the synthesized video (Fig.~\ref{fig:intro}~(d)). While this can yield longer video–action pairs for robot learning, it introduces inefficiencies, degrades cross-modal consistency, and results in large cumulative errors.

To address these issues, we propose VAG, a novel dual-stream generative framework designed to synthesize aligned video–action pairs conditioned on both visual observations and textual prompts (Fig.~\ref{fig:intro}~(e)). Based on flow matching~\cite{lipman2022flow}, it integrates video generation and action generation into a unified dual-stream architecture. The two branches denoise synchronously, enabling the model to produce temporally coherent video sequences alongside semantically consistent action trajectories. Notably, an adaptive 3D pooling module as the bridge between video and action generation has been adopted, which compresses the video latent into a compact global embedding that conditions the action branch. By jointly modeling the video and the action, VAG ensures that the generated visual and motor signals are rigorously aligned at every timestep without cumulative errors, which is crucial for downstream policy learning. As shown in Fig.~\ref{fig:teaser}, VAG forms a complete train-to-deploy pipeline: it learns from teleoperated trajectories, generates aligned video-action pairs from an initial frame and instruction, and supports both policy data synthesis and real-world action replay.

The experimental results suggest that VAG provides meaningful benefits in both generation quality and downstream use. The synthesized trajectories can be executed in simulation and replayed on a real robot with reasonable video-action consistency. We also observe improved action prediction compared with two-stage baselines on both real and simulated datasets. In addition, when VAG-generated data is used for pretraining, the downstream VLA success rate increases from 35\% to 55\% (+20\% absolute). These findings indicate that VAG can serve as a useful data source for improving policy generalization, while leaving room for further improvement in broader settings.

The contributions are summarized as follows: 

\begin{itemize}
\item We propose a novel dual-stream video-action generation framework within a unified Flow-Matching-based generative formulation. Conditioned on the image and textual instruction, it is capable of synthesizing high-quality aligned video-action pairs in a single feed-forward pass.
\item We propose to map the clean latent of the video generation model predicted in every denoising step into an embedding via adapative 3D pooling, effectively compacting the visual information and guiding the action generation.
\item We conduct extensive experiments on both simulated and real-world datasets, demonstrating that our method outperforms other counterparts in terms of video and action prediction. Remarkably, it not only enhances the generalization of policies with its synthesized data, but also holds the potential to function as a world-action policy.
\end{itemize}
\section{Related Work}
\label{sec:related_work}

\textbf{Vision--Language--Action Model.} The prediction of future robot actions from current states and observations has become a core paradigm in robot manipulation. Existing approaches generally fall into two categories. On the one hand, policy-centric models or Vision-Action (VA) models~\cite{chi2025diffusion, zhao2023learningfinegrainedbimanualmanipulation,octo_2023} directly map visual observations to short-horizon actions through diffusion models or behavior cloning. On the other hand, a substantial line of research focuses on Vision-Language-Action (VLA) models~\cite{kim24openvla, tian2025pdfactor,kim2025fine, liu2024rdt, pi_0, pertsch2025fastefficientactiontokenization, intelligence2025pi05visionlanguageactionmodelopenworld, nvidia2025gr00tn1openfoundation, cheang2025gr3technicalreport, zhai2025igniting, jiang2025galaxea, shukor2025smolvla, zheng2025x, li2025mimicdreamer, lin2025onetwovla, li2025cogvla, ni2026swiftvla, team2025gigabrain, wang2025flowram, yu2026dm0, intelligence2025pi, team2026gigabrain, wu2026pragmatic} based on vision-language models (VLM)~\cite{chen2025visrl,chen2025sifthinker,qwen2,paligemma, steiner2024paligemma}. They operate in an iterative closed-loop manner: predicting 1$\sim$2 seconds of actions, executing them, updating the states, and predicting actions again with new states and observations. 

\noindent
\textbf{World Model for Data Synthesis.} World Models~\cite{lu2024drivingrecon,ding2025neural,drivedreamer4d,wang2025sampo,yang2026neoverse} can expand the diversity and scale of visual data that are crucial for downstream robotic skill learning. Recent World Models (WMs), such as SVD~\cite{blattmann2023stable}, Cosmos~\cite{agarwal2025cosmos}, Veo3~\cite{wiedemer2025video}, Wan2.2~\cite{wan2025},  WoW~\cite{chi2025wowworldomniscientworld}, and RoboTransfer~\cite{liu2025robotransfergeometryconsistentvideodiffusion}, generate rich visual rollouts spanning diverse scenarios. Ctrl-World~\cite{guo2025ctrl}, Dreamer4~\cite{hafner2025training},  Veo-Robotics~\cite{team2025evaluating}, DreamDojo~\cite{gao2026dreamdojo} and PlayWorld~\cite{yin2026playworld} treat video prediction as a differentiable simulator by generating videos conditioned on predicted action trajectories. However, video generation alone cannot directly support effective policy learning due to the absence of paired and long-horizon action trajectories. To this end, current methods compensate by relying on externally provided trajectories as additional supervision signals during training.

\noindent
\textbf{World-Action Model as Policy.}
To enhance action prediction, a complementary research direction incorporates future video generation as an auxiliary signal. GR1~\cite{wu2023unleashing}, GR2~\cite{cheang2024gr2generativevideolanguageactionmodel}, WorldVLA~\cite{cen2025worldvla}, UVA~\cite{li2025unified}, DUST~\cite{won2025dualstreamdiffusionworldmodelaugmented}, DreamZero~\cite{ye2026world}, Motus~\cite{bi2025motus}, Cosmos-Policy~\cite{kim2026cosmos}, GigaWorld-Policy~\cite{ye2026gigaworld} and Fast-WAM~\cite{yuan2026fast} jointly predict next-step or multi-frame observations alongside actions. However, these methods primarily focus on improving action prediction via predictive visual signals rather than scalable video-action pair synthesis.

\noindent
\textbf{World-Action Model for Data Synthesis.}
Generating high-quality video–action pairs at scale is increasingly recognized as a key bottleneck for training general-purpose embodied agents. DreamGen \cite{jang2025dreamgen} represents a seminal attempt to synthesize data for robot learning through World Models in a two-stage paradigm. It uses a world model to generate videos and further extracts actions from the generated videos using methods such as IDM (Inverse Dynamics Model)~\cite{baker2022video}. Similarly, AnyPos~\cite{tan2025anypos} regresses actions from generated videos using a vision transformer. Yet the multi-stage, heterogeneous, and asynchronous architecture introduces inefficiencies and degrades cross-modal consistency.

In this work, we address these limitations by introducing VAG, a unified, dual-stream framework that synchronously and efficiently generates both the video and action within a single feed-forward process. VAG produces high-fidelity, coherent video–action pairs approaching 10 seconds, outperforming prior methods in horizon length, consistency, and usability for training downstream robot policies. Experiments on both real-world and simulated datasets confirm that VAG is capable of serving as an effective engine for world-model–driven robot data generation.
\section{Method}

In this work, we focus on joint video-action generation for embodied data synthesis. Leveraging the power of current generative modeling techniques in Sec.~\ref{sec:preliminary}, we propose a novel dual-stream framework, termed VAG, that generates the video and action simultaneously within a unified generative formulation in Sec.~\ref{sec:inference}. Meanwhile, we efficiently train VAG using embodied video-action pairs in Sec.~\ref{sec:training}. 

\begin{figure*}[t!]
    	\centering
    	\includegraphics[width=\textwidth]{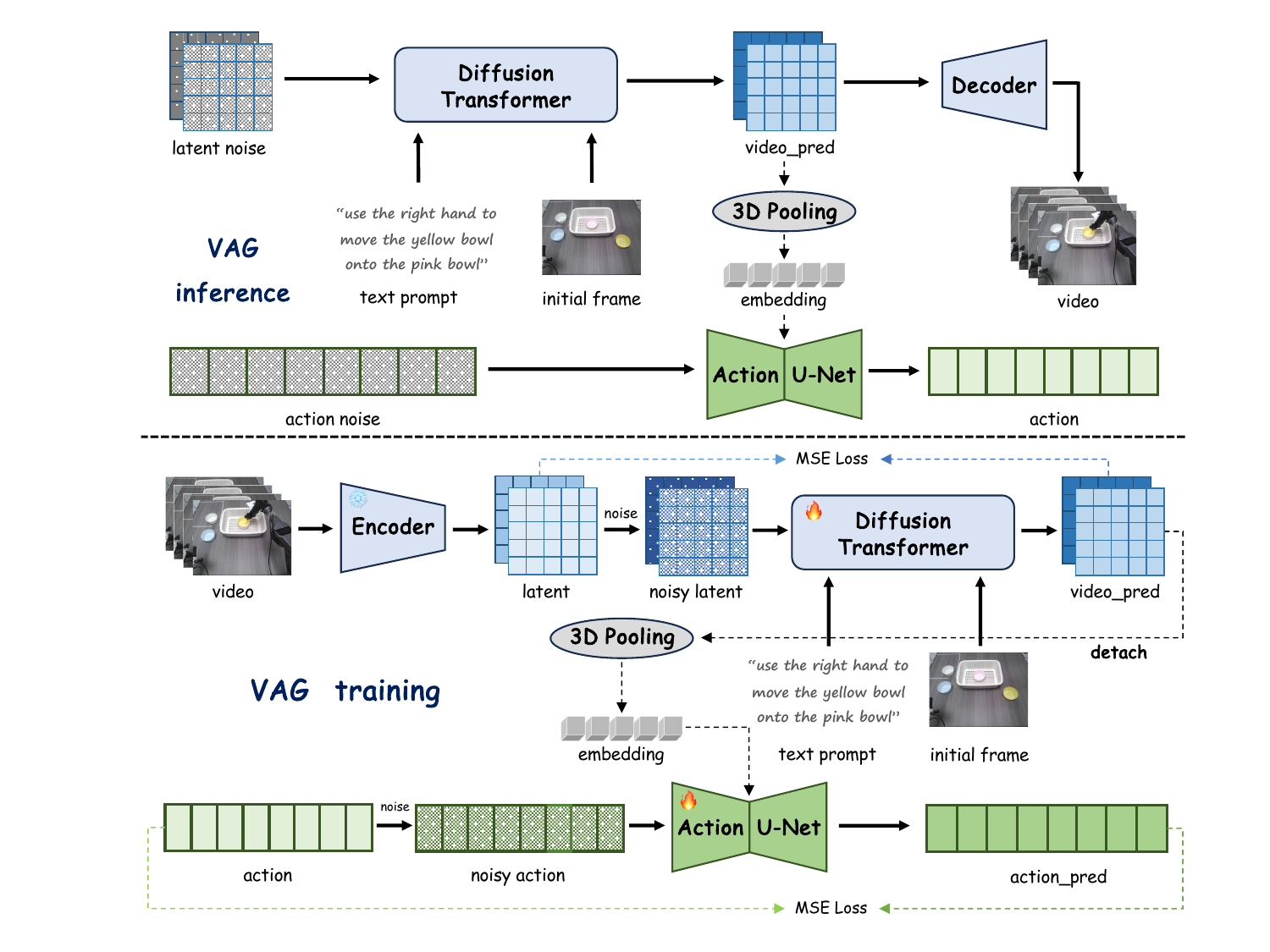}

    	\caption{\textbf{Inference and training pipeline of VAG.} Both the video branch and the action branch are based on flow matching.}
    \label{fig:pipeline}
\end{figure*}

\subsection{Preliminary: Flow Matching}
\label{sec:preliminary}

As a velocity-based formulation, flow matching~\cite{lipman2022flow} not only provides a more direct training target but also tends to
yield smoother optimization and improved sample quality in practice. Formally, given a data sample $\mathbf{x}$, a noise vector $\boldsymbol{\epsilon} \sim \mathcal{N}(0, I)$, and a timestep $t \in[0,1]$ drawn
from a logit-normal distribution, the interpolated latent $\mathbf{x}_{t}$ is defined as:
\begin{align}
\mathbf{x}_t=(1-t) \mathbf{x}+t \boldsymbol{\epsilon}\,.
\end{align}
The corresponding ground-truth velocity is as follows:
\begin{align}
\mathbf{v}_t=\boldsymbol{\epsilon}-\mathbf{x}\,.
\end{align}
The denoising model is trained to predict $\mathbf{v}_t$ by minimizing the mean squared error (MSE) between the prediction and the
ground truth:
\begin{align}
    \mathcal{L}(\theta)=\left\|\mathbf{u}\left(\mathbf{x}_t, t, \mathbf{c} ; \theta\right)-\mathbf{v}_t\right\|^2 ,
\end{align}
where \textbf{c} denotes conditioning information associated with $\mathbf{x}$ (e.g., text embeddings, reference image frames, and
other conditional inputs), $\theta$ represents the model parameters, and $\mathbf{u}(\cdot ; \theta)$ is the predicted velocity function.

\subsection{Dual-Stream Video-Action Generation}
\label{sec:inference}
VAG employs a dual-stream architecture with parallel video and action branches. Conditioned on the image and the textual instruction, it jointly predicts the video and action for $T$ frames. Both streams share the same flow-matching-based formulation to perform synchronized denoising, as shown in the upper subfigure of Fig.~\ref{fig:pipeline}.

\noindent
\textbf{Video Prediction.}
Inherited from the video foundation model Cosmos-Predict2~\cite{agarwal2025cosmos}, the video branch conditions on the image and the text prompt to generate the video in the future. Specifically, to predict a video $\mathbf{V} \in \mathbb{R}^{C \times T \times H \times W}$, a Gaussian noise $\boldsymbol{\epsilon}_{v} \in \mathbb{R}^{C^{\prime} \times (\frac{T-1}{4}+1) \times \lfloor\frac{H}{8}\rfloor \times \lfloor\frac{W}{8}\rfloor}$ is first amplified by the weight $\sigma$ to align with the noise distribution during training. This amplified noise is then concatenated with the input image in latent space, which serves as the prefix condition, and passed through a diffusion transformer (DiT) for denoising. Note that the textual instruction is encoded by T5-XXL~\cite{raffel2020exploring} and added to the DiT through cross-attention layers. To enhance text-context alignment, we adopt classifier-free guidance~\cite{ho2022classifier}. By denoising $\boldsymbol{\epsilon}_{v}$ at each step, the clean latent $\mathbf{z}_{0} \in \mathbb{R}^{C^{\prime} \times (\frac{T-1}{4}+1) \times \lfloor\frac{H}{8}\rfloor \times \lfloor\frac{W}{8}\rfloor}$ under the current noise intensity can be predicted. After $N$ steps of denoising, the final clean video latent is obtained, which is ultimately decoded to yield the predicted video $\mathbf{V}$.

\noindent
\textbf{Action Prediction.}
Alongside the denoising process in the video branch at each step, the action branch receives the clean latent $\mathbf{z}_{0} \in \mathbb{R}^{C^{\prime} \times (\frac{T-1}{4}+1) \times \lfloor\frac{H}{8}\rfloor \times \lfloor\frac{W}{8}\rfloor}$ from the video branch as a condition to guide the action generation. Specifically, to predict the action $\mathbf{A} \in \mathbb{R}^{T \times D}$, a Gaussian noise $\boldsymbol{\epsilon}_{a} \in \mathbb{R}^{T \times D}$ is first initialized. We map $\mathbf{z}_{0}$ to $\mathbb{R}^{C^{\prime} \times 1 \times 1 \times 1}$ using adaptive 3D pooling which averages the entire spatiotemporal features of each channel of $\mathbf{z}_{0}$, and then reshape it into $\mathbb{R}^{1 \times C^{\prime}}$. The vector is then repeated across channels to obtain an embedding $\mathbf{e} \in \mathbb{R}^{1 \times C^{\prime\prime}}$ as a global condition for action generation. This non-learnable approach is simple but efficient, avoiding the need for additional linear layers or complex operations, while preserving global information to ensure that the generated action is rigorously aligned with the generated video. We modify the 1D U-Net from Diffusion Policy~\cite{chi2025diffusion} as the denoiser. Concatenated with the encoded timestep $t$, the embedding $\mathbf{e}$ is fed into U-Net for denoising. After $N$ steps of denoising synchronized with the video branch, action $\mathbf{A}$ is ultimately generated.

\subsection{Training Dual-Stream Video-Action Model}
\label{sec:training}
In this section, we detail the training procedure of VAG. During training, VAG uses embodied video–action pairs and textual instructions describing the robot’s behavior. For each ground-truth video with $T$ frames, we use Qwen2.5-VL~\cite{bai2025qwen2} to extract a textual instruction and encode it with T5-XXL to obtain the corresponding conditioning embedding. In addition, we initialize VAG with the pretrained weights of the underlying video generation model to leverage strong visual priors. Next, we describe the training procedures for the video and action branches in detail.

\noindent
\textbf{Video Branch.}
Given a raw video $\mathbf{V} \in \mathbb{R}^{C \times T \times H \times W}$, we first adopt a VAE~\cite{kingma2013auto} formulation as the visual tokenizer to compress the video with a compression rate of
4 $\times$ 8 $\times$ 8 across the time, height, and width dimensions, respectively. This compression greatly reduces computational cost while preserving essential spatiotemporal structure. After the tokenization, we obtain the latent representation $\mathbf{z} \in \mathbb{R}^{C^{\prime} \times (\frac{T-1}{4}+1) \times \lfloor\frac{H}{8}\rfloor \times \lfloor\frac{W}{8}\rfloor}$ of the video, which is then noise-perturbed to acquire ${\mathbf{z}}^{\prime}$. Conditioned on the first frame of $\mathbf{V}$ and the corresponding textual instruction, we train the DiT to denoise ${\mathbf{z}}^{\prime}$ into ${\mathbf{z}}$ via MSE loss:
\begin{equation}
    \mathcal{L}({\theta}_1)=\left\| \phi_{1}\left(\mathbf{D}\left({\mathbf{z}}^{\prime} ; {\theta}_1\right)\right)-{\mathbf{z}}\right\|^2,
\end{equation}
where $\mathbf{D}$ and ${\theta}_1$ denote the DiT and its parameters. Note that $\phi_{1}(\cdot)$ represents the process of reconstructing the clean latent from the noisy counterpart based on the output of DiT.

\noindent
\textbf{Action Branch.}
After obtaining the predicted clean video latent $\phi_{1}\left(\mathbf{D}\left({\mathbf{z}}^{\prime} ; {\theta}_1\right)\right)$, we apply adaptive 3D pooling over its spatiotemporal dimensions to derive a global video embedding, which serves as the conditioning signal for the action branch. Given the action sequence $\mathbf{A} \in \mathbb{R}^{T \times D}$, in each training step we perturb $\mathbf{A}$ with noise of the same intensity as in the video branch, resulting in $\mathbf{A}^{\prime}$. Conditioned on the detached clean video latent $\phi_{1}\left(\mathbf{D}\left({\mathbf{z}}^{\prime} ; {\theta}_1\right)\right)$, we train a 1D U-Net to denoise $\mathbf{A}^{\prime}$ into $\mathbf{A}$ via an MSE loss:
\begin{equation}
    \mathcal{L}({\theta}_2)=\left\| \phi_{2}\left(\mathbf{U}\left({\mathbf{A}}^{\prime} ; {\theta}_2\right)\right)-{\mathbf{A}}\right\|^2 \ ,
\end{equation}
where $\mathbf{U}$ and ${\theta}_2$ denote the U-Net and its parameters. $\phi_{2}(\cdot)$ represents the process of reconstructing the clean action from the noisy counterpart based on the output of U-Net.
\section{Experiments}

\subsection{Experimental Setup}
\textbf{Implementation Details.}
Our video model is post-trained on Cosmos-Predict2 (2B-Video2World)~\cite{agarwal2025cosmos}, which produces 480P videos with 10 Hz. The number $T$ of video-action frames to be predicted is 93, which means our framework can generate video for approximately the next 10 seconds. Before being fed into the video model, the raw video is automatically resized, with the height $H$ of 432 and the width $W$ of 768. We focus on the RGB video, thus $C$ is 3 and the high-dimensional channel size $C^{\prime}$ is set to 16 according to~\cite{agarwal2025cosmos}. As the global condition for action generation, the length $C^{\prime\prime}$ of the embedding is set to 132 according to~\cite{chi2025diffusion}. During inference, we perform 35 denoising steps ($N$). The training is conducted for 40,000 iterations using 8 NVIDIA H20 GPUs, with a batch size of 1 per GPU.

\noindent
\textbf{Datasets.}
We conduct extensive experiments on two public datasets, including the AgiBot dataset~\cite{bu2025agibot} and the LIBERO dataset~\cite{liu2023libero}, and one self-collected dataset. The AgiBot dataset is a large-scale real-world robotic manipulation dataset, comprising 1 million trajectories across 217 tasks in five deployment scenarios, achieving an order-of-magnitude increase in data scale compared to existing datasets. We only use data collected from the AgiBot G1 dual-arm humanoid robot, with 1794 video-action pairs for training and 200 for testing, where the action dimension $D$ is 16. The LIBERO dataset features a novel simulated dataset across different goals, objects, and layouts. From the subsets LIBERO-Spatial, LIBERO-Object, LIBERO-Goal, and LIBERO-Long, we choose 400 video-action pairs for training and 50 for testing. The action dimension $D$ is 7 because a single-arm robot is simulated. Our self-collected dataset has been curated based on an Agilex Cobot Magic dual-arm robot, with the action dimension $D$ of 14. We allocate 131 samples for VAG training and 20 samples for VLA training, respectively. For the AgiBot dataset and our self-collected dataset, we utilize videos from the head camera. As for the LIBERO dataset, we use stitched videos from both the head and wrist cameras. 

\noindent
\textbf{Baselines.}
We first evaluate the video prediction performance against the state-of-the-art image-to-video models SVD~\cite{blattmann2023stable} and Wan2.2~\cite{wan2025}, which have model sizes of 1.5B and 5B, respectively. Furthermore, to evaluate the accuracy of action prediction, we build two pipelines, using ResNet~\cite{he2016deep} and AnyPos~\cite{tan2025anypos} respectively, to extract action from the video predicted by VAG. For the former, we adopt ResNet50 and MLPs which are commonly used in previous IDM~\cite{baker2022video} works. As for the latter, AnyPos utilizes vision transformer for image-to-action regression. Consistent with VAG, both settings are trained for 40,000 iterations. Additionally, we experiment with the VLA model $\pi_{0.5}$~\cite{intelligence2025pi05visionlanguageactionmodelopenworld}, and investigate the effectiveness and value of VAG in embodied data augmentation via training the VLA with synthetic data generated by VAG.

\begin{figure}[t!]
    \centering
    \includegraphics[width=\linewidth]{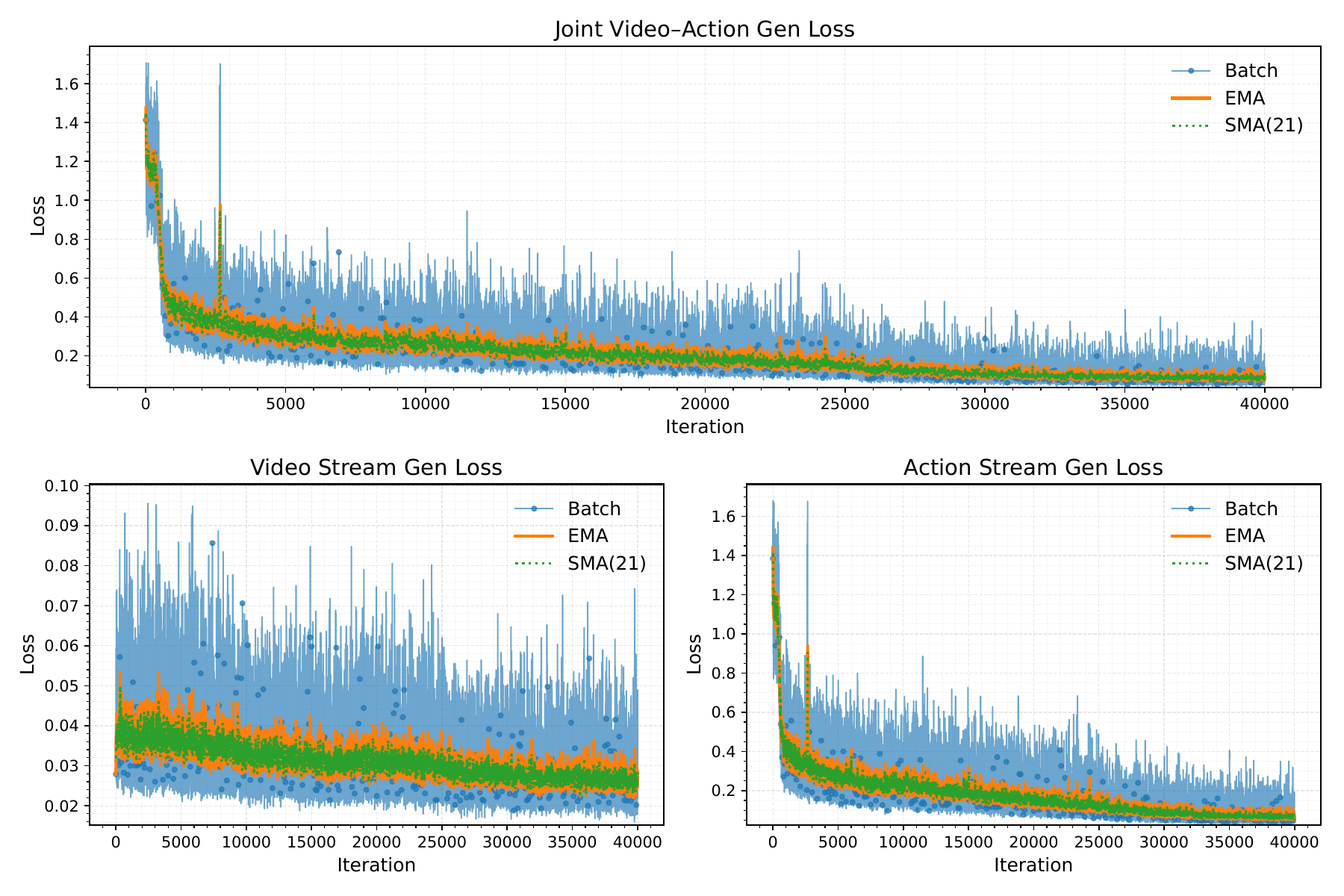}
    \caption{\textbf{Training loss curve of VAG over time on the AgiBot dataset for 40,000 iterations.}}
    \label{fig:training_loss_agibot}
\end{figure}

\begin{figure}[t]
    \centering
    \includegraphics[width=\linewidth]{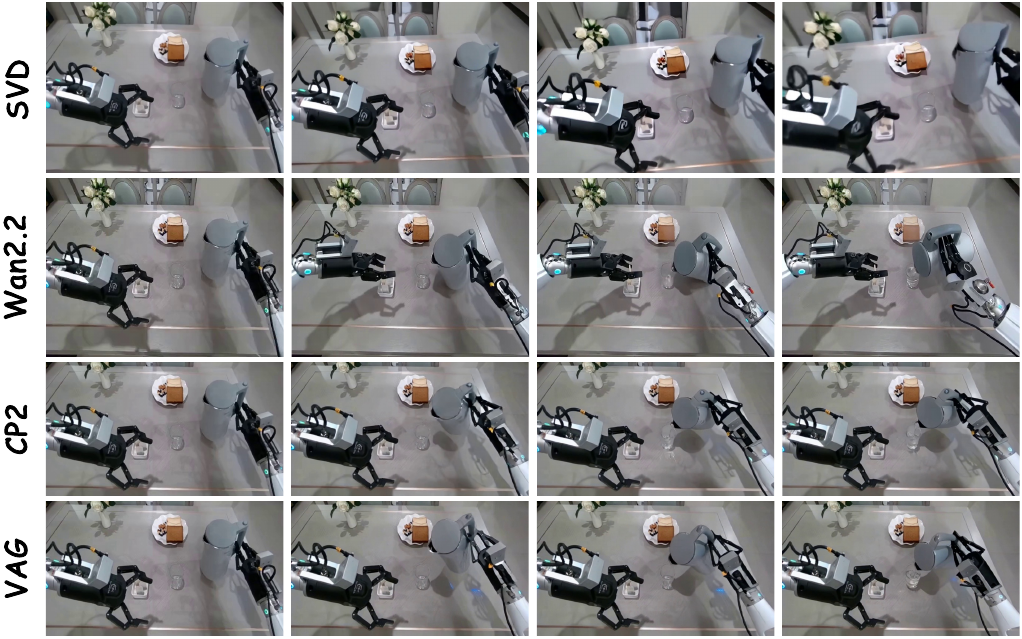}
    \caption{\textbf{Qualitative video generation results across different methods.} Prompt:\textit{“Use the right hand to pour the water from the gray teapot into the cup."} CP2 stands for Cosmos-Predict2.
    }
    \label{fig:videogen}
\end{figure}

\begin{table}[t]
\centering
\resizebox{0.95\linewidth}{!}{
\begin{tabular}{lccccc}
\toprule
Methods & FVD$\downarrow$ & FID$\downarrow$ & LPIPS$\downarrow$ & SSIM$\uparrow$ & PSNR$\uparrow$ \\
\midrule
SVD &    1311 & 150 & 0.421 & 0.339 & 12.7 \\
Wan2.2 & 1152 & \textbf{129} & 0.325 & \textbf{0.612}  & 14.5 \\
CP2 &    988  & 135 & 0.352 & 0.427 & 14.2 \\
VAG &    \textbf{965}  & 130 & \textbf{0.320} & 0.512 & \textbf{15.1} \\
\bottomrule
\end{tabular}
}
\caption{\textbf{Quantitative video generation results on the AgiBot dataset.} CP2 stands for Cosmos-Predict2.}
\label{tab:video_metric}
\end{table}

\begin{table}[t]
\centering
\resizebox{0.95\linewidth}{!}{
\begin{tabular}{lcccc}
\toprule
\multirow{2}{*}{Methods} & \multicolumn{2}{c}{AgiBot} & \multicolumn{2}{c}{LIBERO} \\
\cmidrule(lr){2-3} \cmidrule(lr){4-5}
& ED $\downarrow$ & SR $\uparrow$ & ED $\downarrow$ & SR $\uparrow$ \\
\midrule
VAG-Video + ResNet & 1.54 & 8\% & 0.87 & 37\% \\
VAG-Video + AnyPos & 0.98 & 29\% & 0.55 & 66\% \\
VAG & \textbf{0.81} & \textbf{45\%} & \textbf{0.38} & \textbf{79\%} \\
\bottomrule
\end{tabular}
}
\caption{\textbf{Quantitative action generation results on the AgiBot dataset and LIBERO dataset.} ED stands for Euclidean Distance and SR stands for Success Rate.}
\label{tab:action_metric}
\end{table}

\begin{figure}[t!]
    \centering
    \includegraphics[width=\linewidth]{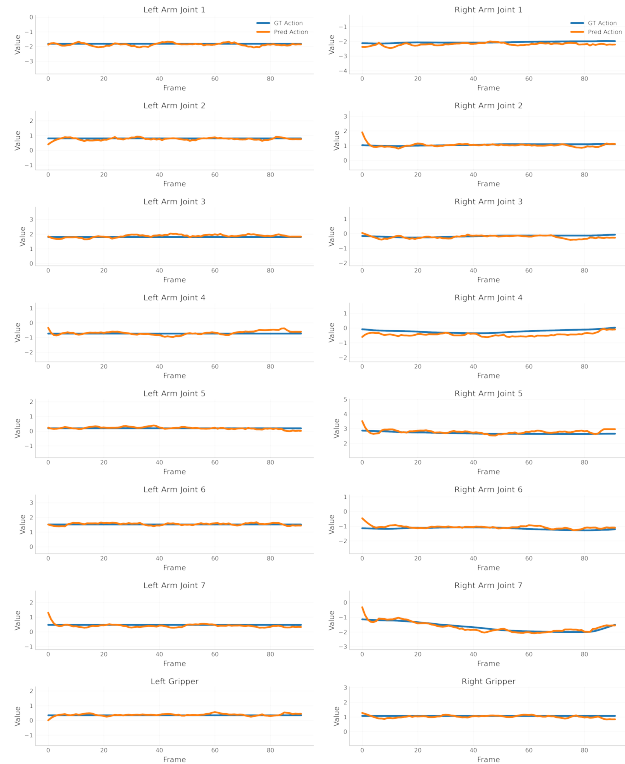}
    \caption{\textbf{Action curve of all 16 dimensions (predicted vs GT) over time on a grasping task of the AgiBot dataset.} Prompt: \textit{“Use the right hand to pick up the apple from the table".}}
    \label{fig:action}
\end{figure}

\subsection{Evaluation in Real-World Environment}

We train VAG on the AgiBot dataset, with the training loss curve shown in Fig.~\ref{fig:training_loss_agibot}, illustrating the optimization dynamics. For video generation, we use the first frame and corresponding text prompt of each test sample. As shown in Fig.~\ref{fig:videogen}, VAG produces videos well aligned with the instructions, demonstrating strong prompt understanding and coherent synthesis. Quantitative results in Tab.~\ref{tab:video_metric} show that VAG achieves the best or competitive performance across multiple metrics.

For action prediction, we evaluate using Euclidean Distance~\cite{danielsson1980euclidean} and Success Rate to measure both numerical accuracy and task-level correctness. Specifically, we compute the Euclidean Distance between predicted and ground-truth actions for each test sample. A prediction is considered successful if the error in each dimension is below 0.2, from which we compute the overall success rate. As shown in Tab.~\ref{tab:action_metric}, compared to a two-stage approach (video generation followed by action regression), VAG achieves the best performance by jointly generating video and action in a unified framework.

\begin{figure}[t!]
    \centering
    \includegraphics[width=\linewidth]{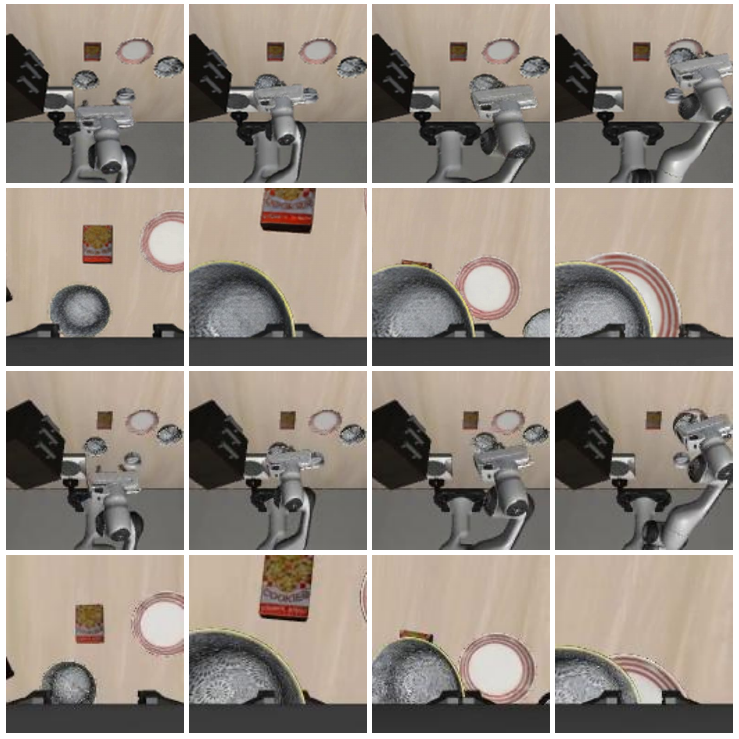}
    \caption{\textbf{Visualization of the generated videos and actions in the simulation environment of LIBERO.} Prompt:\textit{“Pick up the black bowl from table center and place it on the plate."} Top to bottom: generated head-view video; synchronized generated hand-view video; trajectory replay visualization from the head-view; trajectory replay visualization from the hand-view.}
    \label{fig:sim_replay}
\end{figure}

Furthermore, Fig.~\ref{fig:action} displays the predicted action and the ground-truth action across all dimensions on a representative grasping task from the AgiBot dataset, where the predicted action is consistently close to the ground truth. These qualitative plots demonstrate that VAG not only captures the general motion trend but also models fine-grained action variations.

\begin{figure}[t!]
    \centering
    \includegraphics[width=\linewidth]{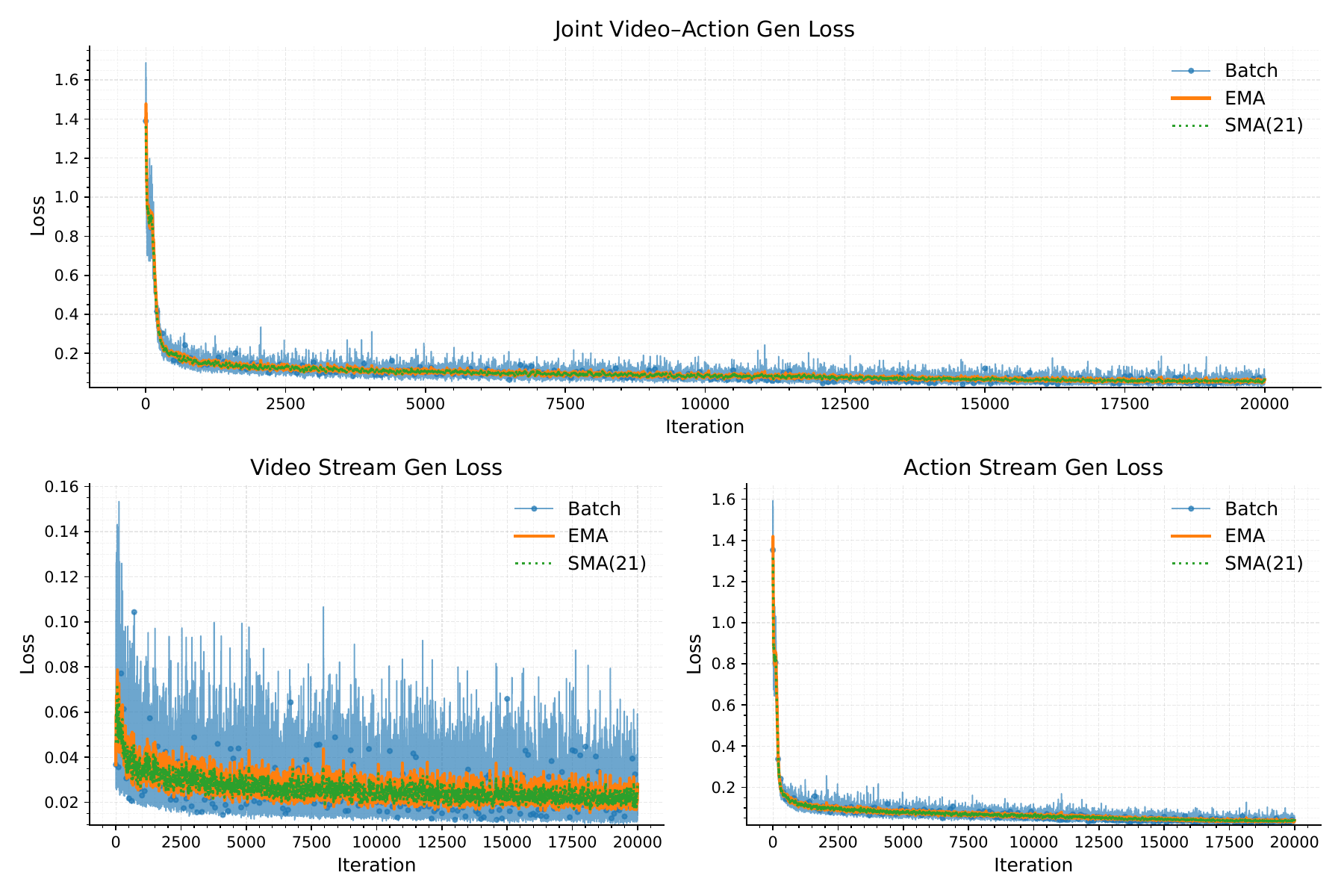}
    \caption{\textbf{Training loss curve of VAG over time on the LIBERO dataset for 20,000 iterations.}}
    \label{fig:training_loss_libero}
\end{figure}

\begin{figure}[t!]
    \centering
    \includegraphics[width=\linewidth]{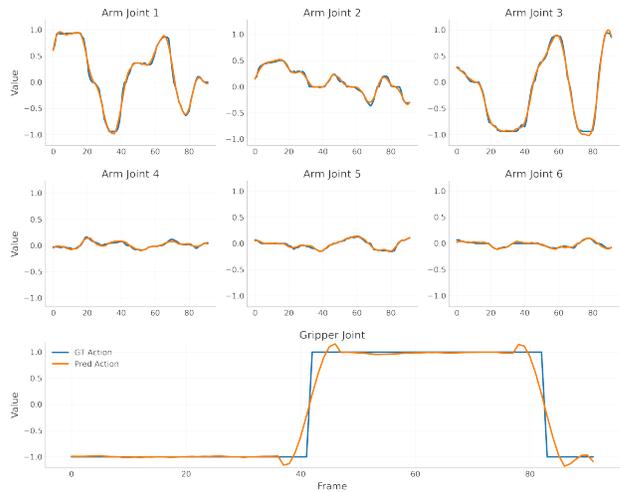}
    \caption{\textbf{Action curves of all 7 dimensions (predicted vs. ground truth) over time on the spatial task of LIBERO dataset.} Prompt: \textit{“Pick up the black bowl between the plate and the ramekin and place it on the plate.”} The blue curves denote ground-truth actions, while the yellow ones denote predicted actions.}
    \label{fig:action_libero}
    \vspace{-3mm}
\end{figure}

\subsection{Evaluation in Simulation Environment}

To further validate the proposed method, we train VAG on the LIBERO dataset for 20,000 iterations, with the training loss curve shown in Fig.~\ref{fig:training_loss_libero}. Compared to AgiBot, LIBERO features more monotonic scenes and action patterns, providing cleaner and more consistent supervision signals during training. As a result, even with only 20,000 iterations, VAG trained on LIBERO shows better convergence, demonstrating stable optimization and strong generalization across different datasets.

We generate videos using the first frame and corresponding text prompt from the test set. As shown in the first two rows of Fig.~\ref{fig:sim_replay}, VAG produces videos well aligned with the given instructions.

\begin{table}[t]
\vspace{1em}
\centering
\resizebox{0.99\linewidth}{!}{
\begin{tabular}{lccccc}
\toprule
\multirow{2}{*}{Methods} & \multicolumn{5}{c}{Success Rate (\%) — Replay } \\
\cmidrule(lr){2-6}
& Spatial & Object & Goal & Long & Avg \\
\midrule
VAG-Video + ResNet & 33 & 34 & 23 & 10 & 25 \\
VAG-Video + AnyPos & 59 & 62 & 56 & 39 & 54 \\
VAG & \textbf{70} & \textbf{72} & \textbf{64} & \textbf{42} & \textbf{62} \\
\bottomrule
\end{tabular}
}
\caption{\textbf{Quantitative action generation results on the LIBERO benchmark.} The best results are marked in \textbf{bold}.}
\label{tab:libreo_action_replay_metric}
\end{table}

For generated trajectories, we evaluate from three aspects: (1) Euclidean distance, (2) trajectory replay, and (3) trajectory visualization. Quantitatively, VAG achieves the lowest Euclidean distance and highest trajectory correctness (Tab.~\ref{tab:action_metric}). In simulation, replayed trajectories successfully complete tasks and closely match the generated videos (last two rows of Fig.~\ref{fig:sim_replay}). We also report success rates in Tab.~\ref{tab:libreo_action_replay_metric}. Visualizations in Fig.~\ref{fig:action_libero} further show that VAG’s predicted trajectories align closely with ground truth in both spatial and temporal aspects.

\subsection{Unlocking Generalization of VLA}

Collecting teleoperation data is labor-intensive and time-consuming. This paper focuses on jointly generating aligned video and action to efficiently synthesize data for policy training. To this end, on our self-collected dataset, we verify the benefit of VAG-synthesized data for VLA. We first train VAG on the training set $\mathcal{X}a$ (131 samples). Then, conditioned on the first frame and corresponding text prompt of each trajectory, we generate aligned video–action pairs, forming a synthetic set $\mathcal{X}{syn}$.

\begin{figure}[t!]
    \centering
    \includegraphics[width=\linewidth]{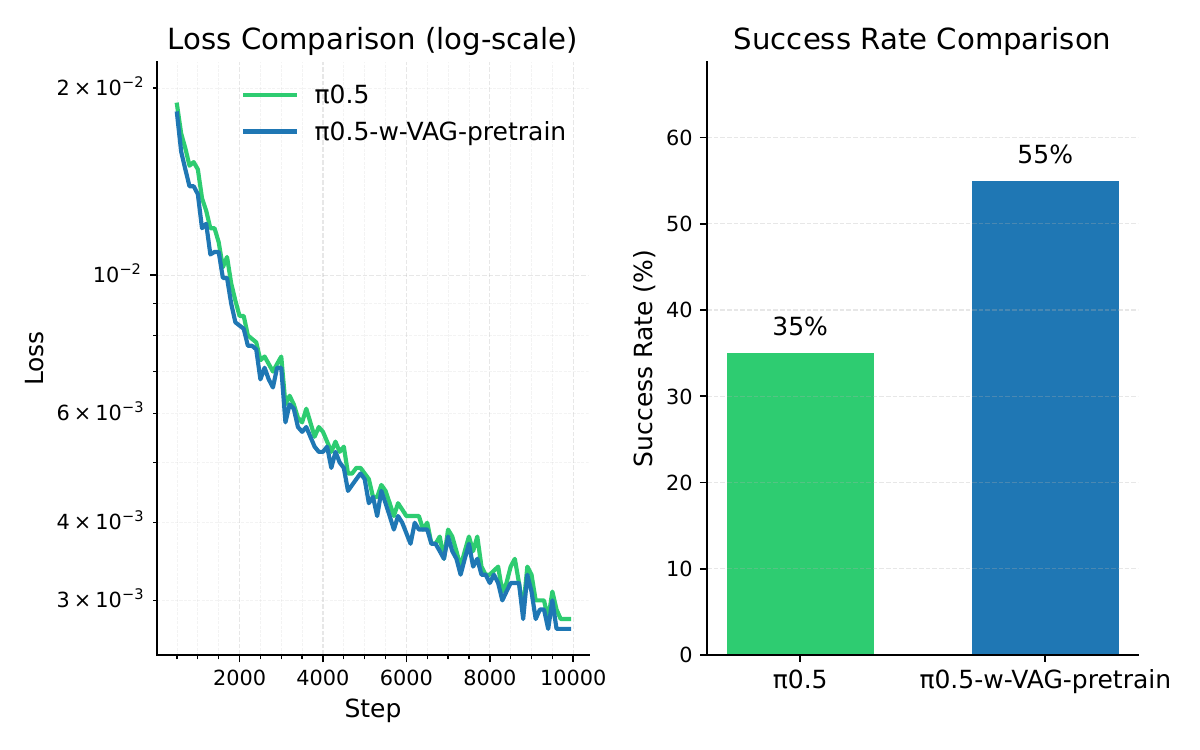}
    \caption{\textbf{Success rate of VLAs in the real world}. Powered by the synthesized data from VAG, $\pi_{0.5}$-w-VAG-pretrain showcases better generalization than $\pi_{0.5}$ (a 20\% increase in success rate).}
    \label{fig:pi05}
    \vspace{-4mm}
\end{figure}

We compare with $\pi_{0.5}$ under two settings. First, $\pi_{0.5}$ is trained on $\mathcal{X}b$ (20 samples) for 10,000 iterations. Second, we pretrain $\pi{0.5}$ on $\mathcal{X}{syn}$ until convergence, then fine-tune on $\mathcal{X}b$ for 10,000 iterations, yielding $\pi{0.5}$-w-VAG-pretrain. We deploy both on a real robot for \textit{pick-and-place} tableware tasks. As shown in Fig.~\ref{fig:pi05}, over 20 trials, $\pi{0.5}$ succeeds 7 times (35\%), while $\pi_{0.5}$-w-VAG-pretrain succeeds 11 times (55\%), achieving a 20\% gain.

\begin{figure}[t!]
    \centering
    \includegraphics[width=\linewidth]{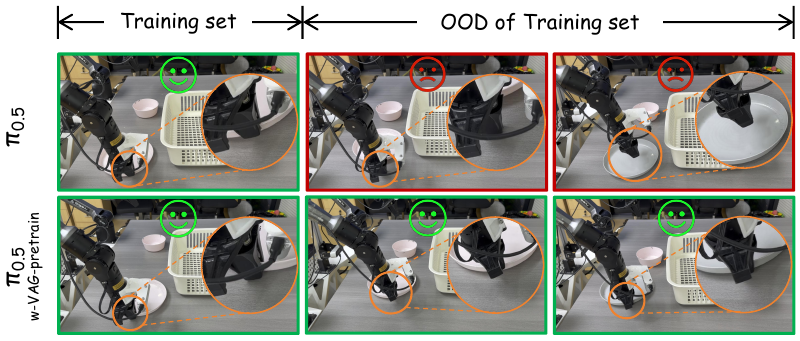}
    \caption{\textbf{Real-world demonstrations of VLAs.} In the face of changes in the position or color of the grasped object, $\pi_{0.5}$-w-VAG-pretrain outperforms $\pi_{0.5}$, verifying VAG's effectiveness.}
    \label{fig:pi05_vs_vag}
    \vspace{-4mm}
\end{figure}

As shown in Fig.~\ref{fig:pi05_vs_vag}, both methods grasp correctly when color and placement appear in $\mathcal{X}b$. However, $\pi{0.5}$ fails under unseen relocation or color changes, while $\pi_{0.5}$-w-VAG-pretrain generalizes better and succeeds consistently. This highlights the effectiveness of VAG-generated data in improving VLA. Notably, although $\pi_{0.5}$-w-VAG-pretrain achieves slightly lower training loss, it does not exhibit the overfitting observed in $\pi_{0.5}$ during real-world deployment.

\subsection{Serving as a World-Action Policy}
\begin{figure*}[t!]
    	\centering
    	\includegraphics[width=0.97\textwidth]{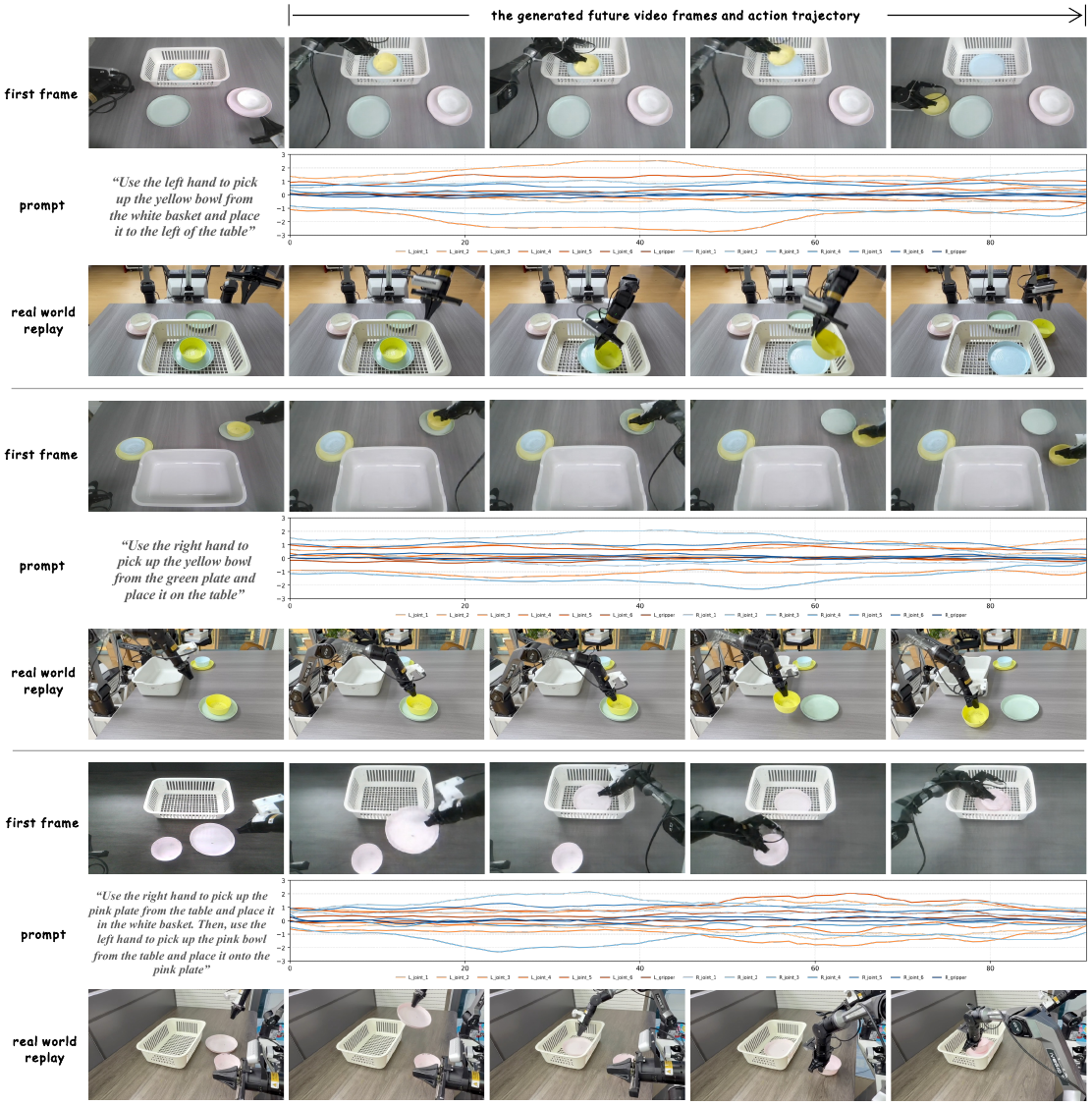}

    	\caption{\textbf{Representative examples of VAG-generated video and action trajectories for left-arm, right-arm, and bimanual manipulation.} Each example includes (1) the initial VAG input frame and generated video, (2) the textual prompt and action trajectory, and (3) the action trajectory executed on the Agilex robot.}
    \label{fig:replay}
    \vspace{-1.5mm}
\end{figure*}

Since VAG can simultaneously generate both video and actions, it essentially serves as a World-Action (WA) model, functioning as a policy. After training VAG on our self-collected dataset, we use the image captured by the head camera of the Agilex robot and the textual instruction as inputs to generate video and actions synchronously. The generated action is then deployed on the Agilex robot for execution. In Fig.~\ref{fig:replay}, we present three representative examples corresponding to left-arm motion and manipulation, right-arm motion and manipulation, and bimanual motion and manipulation, respectively. For each example: (1) The first row shows the initial frame of the input VAG signal in the first column, followed by the generated video frames in subsequent columns; (2) The second row displays the textual prompt in the first column, accompanied by the jointly generated action trajectory in the remaining columns; (3) The third row illustrates the execution of the generated action trajectory replayed on the Agilex robot in the real world. It can be observed that the generated videos and trajectories exhibit high consistency with the real-world robot replay, and the successful execution of the \textit{pick-and-place} task demonstrates VAG’s ability to function as a policy in practical embodied robotic scenarios.
\section{Conclusion}

In this work, we introduce VAG, a generative framework for robot learning with synthetic data. Existing World Models (WMs) and World-Action (WA) models generate realistic videos but lack aligned action trajectories for policy learning, while two-stage approaches introduce inefficiencies and cumulative errors. VAG addresses these by jointly generating videos and actions within a unified flow-matching-based dual-stream framework with synchronized denoising. Experiments show that VAG outperforms existing methods in video-action prediction and improves policy generalization. Overall, VAG provides a scalable approach for embodied data synthesis, reducing reliance on teleoperation and supporting robust visuomotor policy learning.

\noindent
\textbf{Limitations and Future Work.}
VAG currently leverages visual information to guide action generation effectively, whereas video generation has not been influenced by the action branch, wasting beneficial control signals. In the future, we plan to: (1) allow the action branch to guide video generation, further improving the alignment between the generated video and action; (2) replace the action U-Net with DiT for better model capacity; and (3) scale up the training data and conduct experiments on a wider range of tasks.

{
    \small
    \bibliographystyle{ieeenat_fullname}
    \bibliography{main}
}

\end{document}